\documentclass{article}

\usepackage{arxiv}

\usepackage[utf8]{inputenc} 
\usepackage[T1]{fontenc}    
\usepackage[bookmarksnumbered,unicode]{hyperref}
\hypersetup{
    hidelinks
}
\usepackage{xurl}           
\usepackage{url}            
\usepackage{booktabs}       
\usepackage{amsfonts}       
\usepackage{nicefrac}       
\usepackage{microtype}      
\usepackage{lipsum}		   
\usepackage{verbatim}       
\usepackage{graphicx}
\usepackage{natbib}
\usepackage{float}
\usepackage{doi}
\usepackage{siunitx}
\usepackage{algorithmic}
\usepackage{amsmath}
\usepackage{amssymb}
\usepackage{amsthm}
\usepackage[capitalize,noabbrev]{cleveref}
\usepackage{mathtools}
\usepackage{bbold}
\usepackage{enumitem}
\title{Evaluation of Stress Detection as Time Series Events -- A Novel Window-Based F1-metric}

\author{ 
    \href{https://orcid.org/0009-0002-6136-8207}{\includegraphics[scale=0.06]{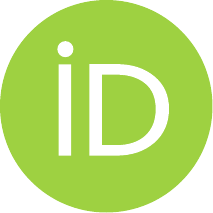}\hspace{1mm}Harald Vilhelm Skat-R{\o}rdam} \\
    Department of Applied Mathematics and\\
    Computer Science\\
    Technical University of Denmark\\
    Denmark\\
    \href{mailto:harsk@dtu.dk}{\texttt{harsk@dtu.dk}} \\
    \And
    \href{https://orcid.org/0000-0002-4017-1280}{\includegraphics[scale=0.06]{orcid.pdf}\hspace{1mm}Sneha Das} \\
    Department of Applied Mathematics and\\
    Computer Science\\
    Technical University of Denmark\\
    Denmark\\
    \href{mailto:sned@dtu.dk}{\texttt{sned@dtu.dk}} \\
    \And
    Kathrine Sofie Rasmussen \\
    Department of Applied Mathematics and\\
    Computer Science\\
    Technical University of Denmark\\
    Denmark\\
    \And
    \href{https://orcid.org/0000-0003-3851-8027}{\includegraphics[scale=0.06]{orcid.pdf}\hspace{1mm}Nicole Nadine L{\o}nfeldt} \\
    Child and Adolescent Mental Health Center\\
    Mental Health Services CPH\\
    Copenhagen University Hospital\\
    Denmark\\
    \href{mailto:nicole.nadine.loenfeldt@regionh.dk}{\texttt{nicole.nadine.loenfeldt@regionh.dk}} \\
    \And
    \href{https://orcid.org/0000-0001-5527-5798}{\includegraphics[scale=0.06]{orcid.pdf}\hspace{1mm}Line Clemmensen} \\
    Department of Mathematical Sciences\\
    University of Copenhagen\\
    Denmark\\
    \href{mailto:lkhc@math.ku.dk}{\texttt{lkhc@math.ku.dk}} \\
}

\date{}

\renewcommand{\shorttitle}{A novel window-based F1-metric}

\hypersetup{
pdftitle={Evaluation of Stress Detection as Time Series Events - A novel window-based F1-metric},
pdfsubject={stat.ME, cs.ML, cs.AI},
pdfauthor={Harald Vilhelm Skat-R{\o}rdam, Sneha Das, Kathrine Sofie Rasmussen, Nicole Nadine L{\o}nfeldt, Line Clemmensen},
pdfkeywords={Time Series, Event Detection, Stress Monitoring, Foundation Models, Wearable Sensors, Reproducibility, Evaluation Metrics},
}

\begin{document}
\maketitle

\begin{abstract}
    Accurate evaluation of event detection in time series is essential for applications such as stress monitoring with wearable devices, where ground truth is typically annotated as single-point events, even though the underlying phenomena are gradual and temporally diffused. Standard metrics like F1 and point-adjusted F1 (F1$_{pa}$) often misrepresent model performance in such real-world, imbalanced datasets. We introduce a window-based F1 metric (F1$_w$) that incorporates temporal tolerance, enabling a more robust assessment of event detection when exact alignment is unrealistic. Empirical analysis in three physiological datasets, two in-the-wild (ADARP, Wrist Angel) and one experimental (ROAD), indicates that F1$_w$ reveals meaningful model performance patterns invisible to conventional metrics, while its window size can be adapted to domain knowledge to avoid overestimation. We show that the choice of evaluation metric strongly influences the interpretation of model performance: using predictions from TimesFM, only our temporally tolerant metrics reveal statistically significant improvements over random and null baselines in the two in-the-wild use cases. This work addresses key gaps in time series evaluation and provides practical guidance for healthcare applications where requirements for temporal precision vary by context.
\end{abstract}


\section{Introduction}\label{submission}
Stress is typically defined as physiological arousal and negative emotion in response to external or internal stimuli~\cite{Lempert2016emotion}. Stress affects our health, well-being, and decision-making capabilities, and our survival is influenced by a well-regulated stress response~\cite{herman2016regulation}. According to the National Institute of Mental Health's (NIMH) Research Domain Criteria, biomarkers of physiological arousal include changes in heart rate, blood pressure, and skin conductance~\cite{NIMH_RDoC_matrix,NIMH_RDoC_Reward_Prediction_Error}. Continuous monitoring of these physiological markers promises to facilitate the development of objective, quantifiable measures of psychiatric symptoms~\citep{Insel} and mental states such as fear, worry, and cognitive effort.
Recent technological advances enable ecologically valid, non-invasive, scalable, and continuous sensing of physiological signals. Wearables and multimodal biosensors are increasingly used for objective stress detection, potentially enhancing the accuracy and reliability of stress assessment \citep{WearablesAndAffectReview}.

Stress and other mental state events have temporal dynamics, including time-to-peak and time-to-return-to-baseline, which depend on the stressor's intensity and modality~\cite{herman2016regulation}. For example, the physiological effects of acute stress extend beyond the initial reaction to the stressor \cite{dickerson2004acute}. However, ground truth labels are often annotated using visible cues rather than physiological signals, and are sometimes described as single time points, leading to a gap between annotation and the underlying phenomena~\cite{Lima2024}.

In stress detection, high precision is important, but it may be equally critical to minimize false discoveries, especially when a prediction may trigger an intervention. For example, in momentary interventions for obsessive-compulsive disorder (OCD), missing an episode may be acceptable if most are detected, but excessive interventions can be disruptive. In contrast, in suicide prevention, missing a stress event could have severe consequences.

Most existing stress datasets are collected in controlled laboratory environments, with relatively few datasets available for in-the-wild stress detection. In real-world settings, data is recorded continuously, but stress events are typically annotated only as isolated time points (point events or point anomalies), in contrast to the continuous stress stimuli of laboratory environments. This annotation paradigm leads to extreme class imbalance, where the proportion of annotated events is negligible compared to the total number of samples. 

To address this, it is common practice to expand point annotations into segments or windows to better reflect the event-like nature of stress episodes (see, e.g., \cite{adarp_dataset,LineandCo}). However, this alters the ground truth labels, resulting in low reproducibility of outcomes across studies and preprocessing pipelines.

The scarcity of annotated events also poses challenges for evaluation. Standard metrics such as F1 and F$_\beta$ become ineffective, as true positives are rare and false positives can be arbitrarily large, resulting in incomplete scoring and unreliable model assessment. Point-adjusted metrics such as F1$_{pa}$~\cite{xu_unsupervised_2018} and F1$_{pa_\%K}$~\cite{kim2022rigorousevaluationtimeseriesanomaly} attempt to address annotation granularity by crediting predictions within labeled event segments, if at least one (or $K\%$) time point is predicted as an event, the entire segment is counted as detected. However, in many real-world stress datasets, only isolated point events are annotated, so such segment-based adjustments are not applicable and do not resolve the underlying class imbalance. Furthermore, as discussed by~\cite{Hand2017}, standard F-metrics and their variants exhibit conceptual weaknesses in balancing the importance of precision and recall.

In this work, we further the investigation into event detection in time series, for three stress-related contexts with strong ecological validity. The ADARP and Wrist Angel datasets were collected \textit{in-the-wild}, meaning participants went about their daily lives and self-annotated stress events as they occurred, while the ROAD dataset was gathered during a structured \textit{real-world} experiment with continuous observer ratings of stress throughout predefined driving sessions.

Our main contribution is a novel evaluation metric that incorporates temporal tolerance, addressing these limitations and enabling more robust assessment of event detection in real-world stress datasets.

\section{Background and Related Work}
In this section, we provide a literature overview of time series prediction, with a focus on models, event detection, and evaluation.

\subsection{Time Series Modeling}
Time series anomaly detection has traditionally relied on statistical models such as ARIMA \citep{hyndman2018forecasting}, which forecast future values and flag deviations as anomalies. Feature-based approaches, including random forests \citep{LineandCo,DataSetNurseStress}, support vector machines, and deep neural networks \citep{Gupta2013}, extract features from time windows for classification tasks. Recent benchmarks indicate that no single algorithm dominates in all anomaly detection setups, and simple methods often perform comparably to more sophisticated ones \citep{AD_TS_Methods}. However, these studies predate the emergence of foundation models for time series.

Since then, foundation models have shown promising results in health-related event detection and classification tasks. Few-shot prompt tuning of large language models (LLMs), significantly improves performance on various health-related tasks, including cardiac, metabolic, physical, and mental health assessments \citep{liu2023largelanguagemodelsfewshot}. The concept has also been extended to physiological signals, such as EEG, with models like the Large Brain Model (LaBraM) outperforming state-of-the-art methods in anomaly detection and event classification \citep{jiang2024largebrainmodellearning}. These advances demonstrate that foundational principles of LLMs can be applied effectively to diverse data modalities, opening new pathways for event detection and classification in healthcare.

According to \citeauthor{ye2024surveytimeseriesfoundation}, only six pre-trained foundation models for time series exist, all trained for general domains. Among these, TimesFM is a decoder-only foundation model pre-trained on over 100B time points, using a patched-decoder architecture that segments time series data into patches for efficient processing and learning from diverse temporal patterns \citep{GoogleTimesFM}. TimesFM demonstrates strong zero-shot forecasting performance across domains and generalizes well due to extensive pre-training on real-world and synthetic time series data.

\subsection{Evaluation of Time-Series Events}
Time series event detection has previously been considered a change point detection problem \citep{Guralnik99, Gensler2018}, without event labels, but where significant changes in the input signal are of interest. This brings an underlying assumption that episodes stretch over time windows and that an event happens in the transition from one such episode to another (change points). In mental state detection, we typically only have specific events labeled, e.g., those related to stress, whereas many other transitions in the physiological time series signal are not considered relevant. Therefore, the event detection problem naturally becomes supervised and labels are needed to indicate which events are relevant.

The standard F$_\beta$ score \citep{chinchor_muc-5_1993} is widely used to balance precision and recall in binary classification, where the $\beta$ parameter adjusts the weight of precision. However, in time-series event detection, this pointwise metric may not fully capture the temporal structure of events.

Recently, \citeauthor{Lima2024} introduced detection probability and detection lag as metrics to address the gap between an event's occurrence and its detection. These metrics acknowledge detections within a given time lag as correct, but do not fully address issues such as false discoveries or class imbalance. When detecting anomalies in time series, it is often more important to detect anomalous events rather than anomalous time points, but this is often not reflected in evaluation metrics \citep{DetectionMetric}.

Similarly, \citeauthor{xu_unsupervised_2018} proposed point adjustment (pa) a method which adjusts the predicted events such that if there is just one time point predicted as an event within a sequence of labeled events, then all time points within the sequence are adjusted with predictions as events before the F1 score is computed. This method requires information about the true labels and thus can only be used for evaluation, not for adjusting predictions of unseen data. According to \citeauthor{kim2022rigorousevaluationtimeseriesanomaly}, the use of F1$_{pa}$ shows a tendency to overestimate, while the classic F1 leans more toward underestimating. Therefore, they introduced a tweak to F1$_{pa}$ with F1$_{pa\%K}$, where at least $K\%$ time point predictions within a sequence are required for adjustment.

We propose an F1-metric, F1$_w$, which takes into account a possible time lag, $w$, providing a temporally tolerant evaluation of model performance, and evaluate its effectiveness on three time series stress event detection datasets.

\section{Methods}

\subsection{Problem Formulation}
The observed time-series signal from $N$ sensors in wearable devices, at time points $t = 1,\ldots, T$, is denoted as $X = \{\mathbf{x}_1,...,\mathbf{x}_T\}$, with $\mathbf{x}_t \in \mathbb{R}^N$. The event label $y_t \in \{0,1\}$ indicates whether a time point is in an event (1) or not (0). Typically, tagged events assign $y_{t} = 1$ for a single time point $t$ without duration, although the underlying ground truth may be more continuous.

\subsection{F1 Metrics}\label{sec:f1_metrics}
\Cref{tab:f1_variants_pa_window} summarizes the definitions of the standard F1, the point-adjusted F1 with a $K\%$ threshold (pa\%K) \citep{kim2022rigorousevaluationtimeseriesanomaly}, and our proposed window-based F1 metric. When $K=0$ for pa\%K, it yields the standard point-adjusted metric \citep{xu_unsupervised_2018}; when $K=1$, it recovers the standard pointwise F1. The window-based F1 metric (F1$_w$) tolerates temporal offsets by considering predictions within a window around each event.

\begin{table}[!tbh]
\centering
\caption{Summary of F1 metric variants and their definitions. $S_k$ denotes the $k$th true event segment. For pa\%K, $K \in [0,1]$ is a threshold; for $K=0$, pa\%K reduces to the standard point-adjustment metric (pa); for $K=1$, pa\%K is equivalent to the standard pointwise metric (F1). For the window-based method, $w_t$ is the window of size $w$ around the time point $t$.}
\label{tab:f1_variants_pa_window}
\begin{tabular}{l|c|c}
\hline
 & \textbf{Point-Adjusted with K\% (pa\%K)} & \textbf{Window-based (w)} \\
\midrule
TP & $\displaystyle\sum_{k}\sum_{t \in S_k} \mathbb{1}\left(\hat{y}_t = 1 \lor \frac{\sum_{t' \in S_k}\mathbb{1}(\hat{y}_{t'} = 1)}{|S_k|} \ge K\right)$ 
   & $\displaystyle\sum_{t: \hat{y}_t=1} \mathbb{1}\left(\exists\, t' \in w_t : y_{t'} = 1\right)$ \\[3ex]
FP & $\displaystyle\sum_{t=1}^T \mathbb{1}(\hat{y}_t = 1 \land y_t = 0)$ 
   & $\displaystyle\sum_{t: \hat{y}_t=1} \mathbb{1}\left(\forall\, t' \in w_t : y_{t'} = 0\right)$ \\[3ex]
FN & $\displaystyle\sum_{k}\sum_{t \in S_k} \mathbb{1}\left(\hat{y}_t = 0 \land \frac{\sum_{t' \in S_k}\mathbb{1}(\hat{y}_{t'} = 1)}{|S_k|} < K\right)$ 
   & $\displaystyle\sum_{t: y_t=1} \mathbb{1}\left(\forall\, t' \in w_t : \hat{y}_{t'} = 0\right)$ \\[3ex]
Precision & $\displaystyle\frac{\text{TP}_{pa\%K}}{\text{TP}_{pa\%K} + \text{FP}}$ 
          & $\displaystyle\frac{\text{TP}_w}{\text{TP}_w + \text{FP}_w}$ \\[3ex]
Recall & $\displaystyle\frac{\text{TP}_{pa\%K}}{\text{TP}_{pa\%K} + \text{FN}_{pa\%K}}$ 
       & $\displaystyle\frac{\text{TP}_w}{\text{TP}_w + \text{FN}_w}$ \\[3ex]
FDR & $\displaystyle\frac{\text{FP}}{\text{TP}_{pa\%K} + \text{FP}}$ 
    & $\displaystyle\frac{\text{FP}_w}{\text{TP}_w + \text{FP}_w}$ \\[3ex]
F-metric & $\displaystyle 2\cdot \frac{\text{Precision}_{pa\%K}\cdot \text{Recall}_{pa\%K}}{\text{Precision}_{pa\%K}+\text{Recall}_{pa\%K}}$ 
   & $\displaystyle 2\cdot \frac{\text{Precision}_w\cdot \text{Recall}_w}{\text{Precision}_w+\text{Recall}_w}$ \\[3ex]
\hline
\end{tabular}
\end{table}

To balance the relative importance of recall and precision, the F$_\beta$ score is defined as $F_\beta = \frac{(1+\beta^2)\cdot\text{Precision} \cdot \text{Recall}}{(\beta^2 \cdot \text{Precision}) + \text{Recall}}$, where $\beta$ controls the weighting; $\beta=1$ yields the standard F1 score (harmonic mean). While this balance can in principle be applied to all F-metrics, in this study it is only used for the standard F1 metric.

\begin{figure}[!htb]
    \centering
    \includegraphics[width=\textwidth]{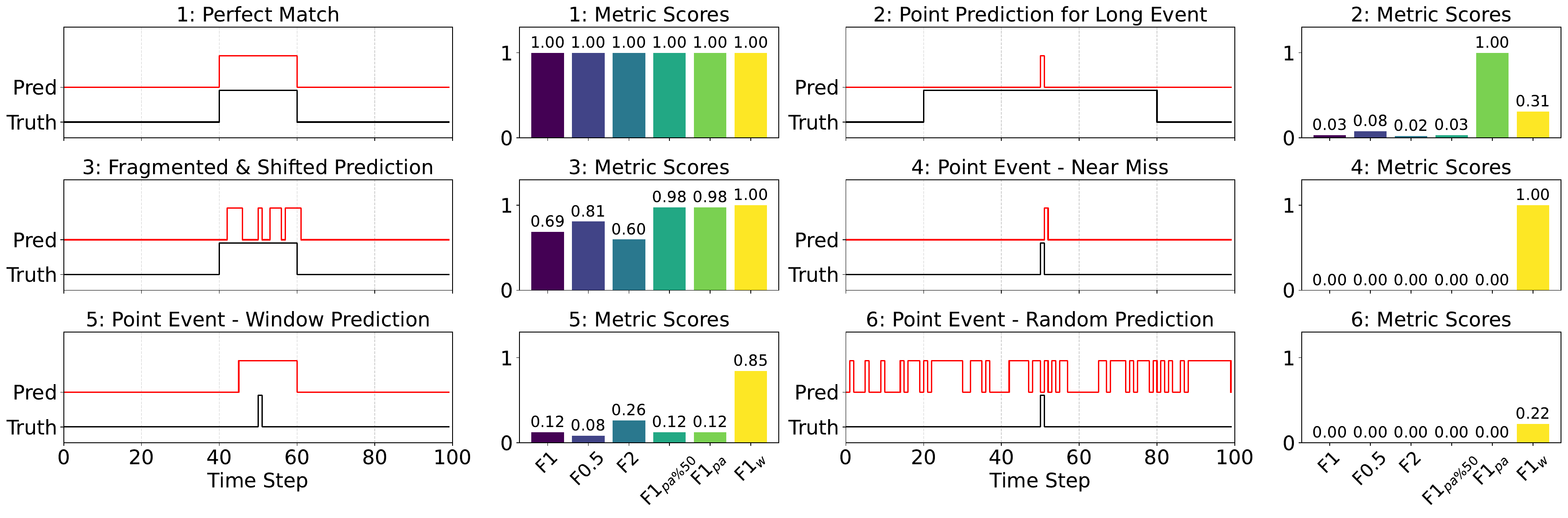}
    \caption{Illustrative scenarios for event prediction evaluation. Each row shows a different synthetic scenario comparing ground truth events ($y$) and model predictions ($\hat{y}$) over time (left), alongside a bar chart of six F1-based metrics (right): standard F1 ($K=1$), F$_\beta$ with $\beta=0.5$ and $\beta=2$, point-adjusted F1 with $K=0.5$ (F1$_{pa\%K}$), point-adjusted F1 (F1$_{pa}$, $K=0$), and window-based F1 (F1$_w$) with a window of $w=10$ time steps.}
    \label{fig:FLA}
\end{figure}

We illustrate the F-based metrics in \Cref{fig:FLA} using six synthetic scenarios:

\begin{itemize}
    \item[(1)] \textbf{Perfect Match:} All F-metrics yield 1.
    \item[(2)] \textbf{Point Prediction for Long Event:} F1$_{pa}$ overestimates (F$_{pa}$ = 1); F1$_{pa\%K}$ and standard F-metrics are low; F1$_w$ gives partial credit.
    \item[(3)] \textbf{Fragmented \& Shifted Prediction:} F1$_w$ achieves a perfect score by tolerating temporal offset; other metrics penalize late or fragmented detection.
    \item[(4)] \textbf{Point Event - Near Miss:} Only F1$_w$ is nonzero, reflecting its tolerance to small misalignments.
    \item[(5)] \textbf{Point Event - Window Prediction:} F1$_w$ gives partial credit for predictions near the event, but penalizes excess window; others score near zero.
    \item[(6)] \textbf{Point Event - Random Prediction:} All metrics are zero except F1$_w$, which is slightly elevated due to the event density in the random predictions.
\end{itemize}

\subsection{Prediction Model}
We perform zero-shot forecasting using TimesFM (\verb|google/timesfm-2.0-500m-pytorch|) with co\-variates. The model uses a context length of 2048 for its main target signal (i.e. stress) (8min 32sec), a prediction horizon of 256 (64sec), and a combined covariate context length of 2304 (2048+256) (see details in \cref{fig:illustration}). Hyperparameters are set as follows: 50 layers, no positional embedding, and per-core batch size of 2400 for GPU and 256 for CPU. TimesFM is designed for continuous univariate time series forecasting, thus to predict events, we apply a sigmoid activation to the output layer. Thresholds for event prediction are selected based on validation results for each dataset (\Cref{fig:ADARP_val_prob_dist,fig:ROAD_val_prob_dist,fig:WA_val_prob_dist}), as shown in \Cref{tab:thresholds}.

\begin{table}[ht]
\centering
\caption{Threshold ($\delta$) values for each dataset.}
\label{tab:thresholds}
\begin{tabular}{lccc}
\toprule
\textbf{Dataset} & \textbf{ADARP} & \textbf{ROAD} & \textbf{WA} \\
\midrule
Threshold ($\delta$) & 0.501 & 0.71 & 0.501 \\
\bottomrule
\end{tabular}
\end{table}

\subsection{Baseline Models}
We compare TimesFM to two baseline models: a random baseline, which assigns event probabilities uniformly at random, and a null baseline, which is set to yield an F-score of 0. Note that this is not the same as assigning 0 probability to all events. These baselines provide reference points for evaluating the predictive performance of TimesFM.

\subsection{Statistical Testing Methods}
For each metric and dataset, we perform subject-level F-metric permutation tests to compare TimesFM scores against both random and null baselines. The combined significance criterion requires that, for a given metric, TimesFM outperforms both baselines at $\alpha=0.05$, only then is significance reported. Confidence intervals for the difference to random baseline are estimated via bootstrap resampling. 

For all statistical tests, F-metric values are rounded to two decimal places before computing differences. This ensures that only meaningful differences are considered and prevents detecting significance for negligible differences (e.g., 0.0001).

\subsubsection{Permutation Test}
Let $x_i$ and $y_i$ denote the F-metrics for subject $i = 1, \dots, n$ for TimesFM and the baseline model, respectively. The per-subject difference is defined as: $d_i = x_i - y_i$, with the mean difference: $\bar{d} = \frac{1}{n} \sum_{i=1}^n d_i.$
Under the null hypothesis the model and metric choice has no effect, the distribution of $\bar{d}$ is symmetric around zero. To approximate the null distribution, we generate $B = \min(10{,}000,\ 2^n)$ sign-flipped versions of the differences by randomly multiplying each $d_i$ by $+1$ or $-1$, and compute the permuted statistic: $\bar{d}^{(b)} = \frac{1}{n} \sum_{i=1}^n s_i^{(b)} d_i, \quad \text{where } s_i^{(b)} \in \{-1, +1\}$.
The two-sided $p$-value is then estimated as:
\begin{equation*}
    p = \frac{1 + \sum_{b=1}^{B} \mathbb{1} \left( \left| \bar{d}^{(b)} \right| \geq \left| \bar{d} \right| \right)}{1 + B}.    
\end{equation*}

\subsubsection{Bootstrap Confidence Intervals}
To estimate uncertainty around the mean difference $\bar{d}$, we apply non-parametric bootstrap resampling. We generate $B = \min(10{,}000,\ \binom{2n-1}{n})$ bootstrap samples by sampling the $n$ subject-level differences $\{d_1, \dots, d_n\}$ with replacement. For each resample $b = 1, \dots, B$, we compute:
$$
\bar{d}^{(b)} = \frac{1}{n} \sum_{i=1}^n d_i^{(b)},
$$
where $d_i^{(b)}$ is the $i$-th sampled difference in bootstrap sample $b$. The 2.5th and 97.5th percentiles of $\{\bar{d}^{(1)}, \dots, \bar{d}^{(B)}\}$ form the 95\% confidence interval for the mean difference.

\section{Data and Experiments}
We evaluate three event-based metrics across three datasets using zero-shot forecasting for event prediction. Two datasets (ADARP and Wrist Angel) capture physiological arousal in mental health contexts, with stress events labeled via self-reported button presses. The third dataset (ROAD) contains physiological data collected during driving under externally induced stress, with stress labels based on observer-rated continuous scores. The self-reported datasets provide only point events, highlighting limitations of current metrics in real-world settings. In contrast, the experimental ROAD dataset enables the use of additional F1 metrics not applicable to the self-reported datasets.

\subsection{Data}
This section describes the three datasets and the physiological signals collected using the Empatica E4 wearable device.

\subsubsection{Data from the E4 wearable:} 
The datasets were all collected using the Empatica E4, a Class IIa medical device~\citep{medical_device_coordination_group_mdcg_2021}, which is ISO 13485 certified, HIPAA compliant, GDPR compliant, and FDA-cleared~\citep{Empatica_Legal}. The E4 is equipped with four sensors: photoplethysmography (PPG, 64Hz, for blood volume pulse/BVP and heart rate/HR at 1Hz), electrodermal activity (EDA, 4Hz), skin temperature (4Hz), and a 3-axis accelerometer (32Hz). The device also features an event mark button for timestamped event tagging. For this study, we used HR, BVP, EDA, and 3-axis accelerometer signals, all resampled to 4Hz.

\begin{itemize}
    \item \textbf{ADARP}: The ADARP dataset~\citep{adarp_dataset} is a real-world study of 11 adults with alcohol use disorder (AUD), each wearing the E4 for 14 days (average 11.5 hours/day). Stress events were self-tagged using the event button when participants felt “more stressed, overwhelmed, or anxious than usual.” The dataset contains 1,698 hours of physiological data and 409 recorded stress events.
    \item \textbf{ROAD}: The ROAD dataset~\citep{Road_dataset} was collected from 14 driving sessions with 10 adults. Each session followed a fixed route with both high-stress urban and low-stress highway segments. Participants wore E4 devices on both wrists, of which only the left wrist data is used. An observer continuously rated driver stress using a slider (0 = no stress, 1 = extreme stress), validated by participants. Stress events were defined using a threshold of 0.75~\citep{Bustos_Elhaouij_Sole}. The dataset contains 10.7 hours of data.
    \item \textbf{Wrist Angel}: The Wrist Angel dataset~\citep{LønfeldtProtocol} is an in-the-wild observational case-control study of nine youths (ages 10–16) diagnosed with OCD. Participants wore the E4 on their non-dominant wrist during waking hours for 4–48 days (total 2,405 hours~\citep{LineandCo}). They pressed the event mark button whenever they felt stressed by OCD symptoms. Accidental tags at session boundaries were removed.
\end{itemize}

\subsection{Data Processing}
Signals were filtered using Butterworth filters~\citep{haykin2007signals}: BVP (2nd order, 2–12Hz band-pass), EDA and temperature (6th order, 1Hz low-pass). All signals were resampled to 4Hz. Data points with HR below 40 bpm (ADARP, ROAD), 50 bpm (Wrist Angel), or temperature outside 23–50$^\circ$C, were excluded~\citep{HRChildren,LineandCo,MoreDataIsGod}. This results in the data distribution shown in \Cref{tab:compact_data_overview}.

\subsubsection{Splitting}
For ADARP and Wrist Angel, a temporal within-subject split was performed using: the first 20\% of each subject’s data for validation, the remaining 80\% for testing. For ROAD, a subject-wise split was performed: 20\% of subjects for validation, 80\% for testing. Data normalization was performed using the validation set. \Cref{tab:compact_data_overview} shows the distribution of subjects and events in each split.

\begin{table}[!tbh]
\centering
\caption{Characteristics and split distribution of the three pre-processed datasets. An event corresponds to a single time point at 4Hz labeled as an event.}
\label{tab:compact_data_overview}
\setlength{\tabcolsep}{3pt}
\begin{tabular}{l r r r r}
\toprule
Dataset    & \#Subjects & \#Events & \#Data Points   & Event (\%) \\
\midrule
\multicolumn{5}{l}{\textbf{ADARP}} \\
\quad All          & 11         & 405      & 24\,020\,694    & \num{0.001686} \\
\quad Validation   & 11         & 89      &  4\,804\,142    & \num{0.002617} \\
\quad Test         & 11         & 319      & 19\,216\,552    & \num{0.001333} \\
\addlinespace
\multicolumn{5}{l}{\textbf{ROAD}} \\
\quad All          & 12         & 92\,783   &    154\,116     & 60                 \\
\quad Validation   &  3         & 28\,177   &     40\,312     & 70                 \\
\quad Test         &  9         & 64\,606   &    113\,804     & 57                 \\
\addlinespace
\multicolumn{5}{l}{\textbf{Wrist Angel}} \\
\quad All          & 8     & 2456   & 39\,976\,356          & \num{0.006143}            \\
\quad Validation   & 8     & 721   & 7\,995\,274           & \num{0.009017}            \\
\quad Test         & 8     & 1735   & 31\,981\,082          & \num{0.005425}            \\
\bottomrule
\end{tabular}
\end{table}

\section{Results}
In this section, we first describe the performance of the metrics for our stress event forecasting with TimesFM. A comprehensive statistical analysis is subsequently presented to evaluate their significance. Details on hardware specifications and inference runtimes for each dataset are provided in Appendix~\ref{sec:runtime_details}.

\subsection{F-Metric Comparison}
\begin{figure}
    \centering
    \includegraphics[width=0.98\linewidth]{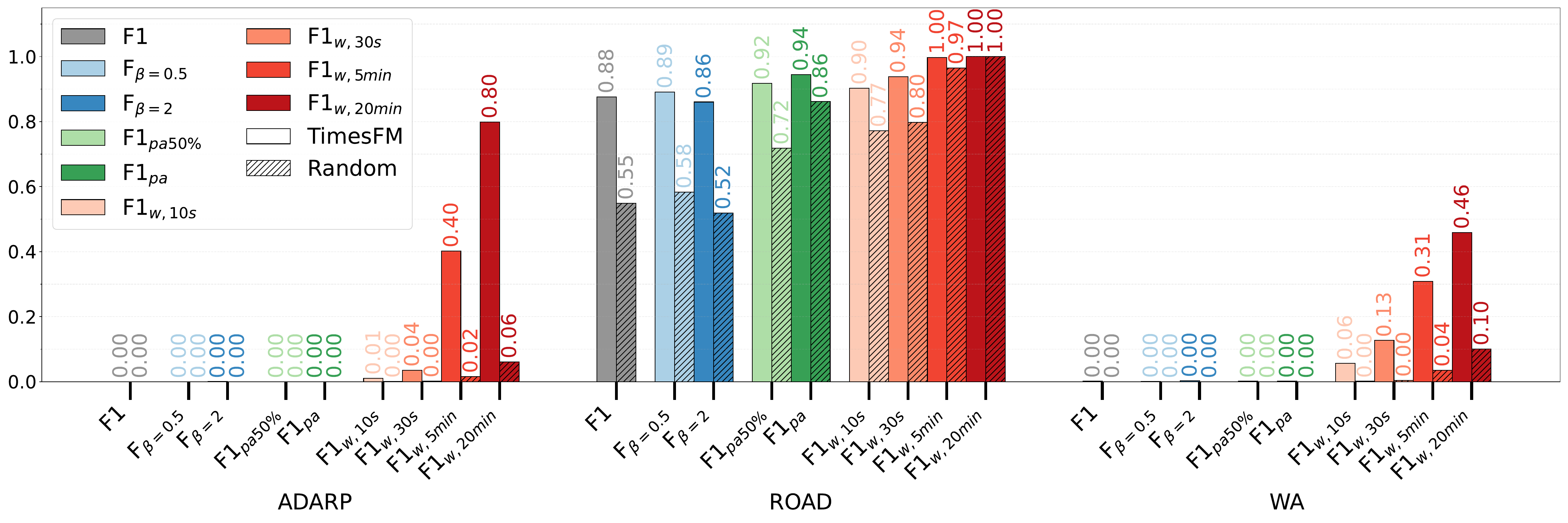}
    \caption{Comparison of TimesFM and random baseline performance across multiple metrics and datasets. Bars show F1 and related scores for standard, point-adjusted, and window-based metrics for ADARP, ROAD, and Wrist Angel. For each metric, TimesFM (solid bars) is compared to a random baseline (hatched bars). Metric values are annotated above each bar for clarity.}
    \label{fig:bar_plot_comp}
\end{figure}

The bar plot in \Cref{fig:bar_plot_comp} illustrates the contrasting impressions of model performance obtained with the different evaluation metrics, depending on their tolerance to temporal offsets and the characteristics of each dataset. 

\begin{itemize}
    \item \textbf{ADARP}: Standard and point-adjusted metrics (F1, F$_{\beta}$, F1$_{pa}$, F1$_{pa50\%}$) report zero scores, suggesting no apparent predictive ability. In contrast, window-based metrics (F1$_w$) reveal increasing scores as the window size grows, indicating that predictions are often temporally close to true events even if not perfectly aligned. The random baseline remains low across all metrics, emphasizing the added value of temporal tolerance.
    \item \textbf{ROAD}: The long event segments in the ROAD data result in a high prevalence of events (57\%), causing all point-adjusted and window-based metrics to yield high scores for the random baseline (all above 0.7), which would typically indicate a useful model for this domain. For our window-based metric, the random baseline score decreases as the window size becomes smaller, naturally approaching the standard point-based metric ($\lim\limits_{w \to 0} \mathrm{F1}_w = \mathrm{F1}$), which still reports a relatively high score of around 0.5. This demonstrates that adjusted metrics cannot be used interchangeably across datasets and must be applied with caution, as they may overestimate performance in cases with dense labeling or large continuous event segments.
    \item \textbf{Wrist Angel}: TimesFM's reported performance is similar to that on ADARP. Standard and point-adjusted metrics yield zero scores. In contrast, the window-based F1$_w$ metric shows model scores that increase with window size: 0.06 (10s), 0.13 (30s), 0.31 (5min), and 0.46 (20min). The random baseline scores remain consistently lower (e.g., 0 for 10s and 30s, 0.04 for 5min, and 0.10 for 20min), suggesting that, as with ADARP, only temporally tolerant evaluation captures the model's predictive ability.
\end{itemize}

\subsection{Statistical Significance}
The results of the subject-level permutation tests and bootstrap confidence intervals are summarized in \Cref{tab:combined_significance}. For each metric, significance is only reported if TimesFM outperforms both the random and null baselines.

While the previous section illustrated how different metrics yield contrasting performance trends and gaps between TimesFM and baselines, this section provides a formal statistical validation of these observations. We use subject-level permutation tests to determine if the model's performance is significantly better than both a random and a zero-score (null) baseline. The results, including 95\% bootstrap confidence intervals for the performance difference, are summarized in \Cref{tab:combined_significance}. Significance is reported only when TimesFM outperforms both baselines.

\begin{table}[ht]
\centering
\caption{Combined significance test: TimesFM must outperform both Random and Null baselines. $p$-value shown is the more conservative (higher) of the two tests. Significance stars ($^{***}P<0.001$, $^{**}P<0.01$, $^{*}P<0.05$) appear only when both tests are significant. Failure indicators: $^{\nsim R}$ not significantly different from random, $^{\nsim 0}$ not significantly different from null. Degenerate cases: $^{\dagger}$ all scores are 0, $^{\ddagger}$ all scores are 1. CI indicates the 95\% confidence intervals for the difference in F-metric means between TimesFM and the random baseline.}
\label{tab:combined_significance}
\begin{tabular}{@{}l*{6}{c}@{}}
\toprule
 & \multicolumn{2}{c}{\textbf{ADARP}} & \multicolumn{2}{c}{\textbf{ROAD}} & \multicolumn{2}{c}{\textbf{WA}} \\
\cmidrule(lr){2-3} \cmidrule(lr){4-5} \cmidrule(lr){6-7}
\textbf{Metric} & $p$-value & 95\% CI & $p$-value & 95\% CI & $p$-value & 95\% CI \\
\midrule
F1 & 1.000$^{\nsim R}$$^{\nsim 0}$ & [0.000, 0.000]$^{\dagger}$ & 0.004$^{**}$ & [0.288, 0.361] & 1.000$^{\nsim R}$$^{\nsim 0}$ & [0.000, 0.000]$^{\dagger}$ \\
F$_{\beta=0.5}$ & 1.000$^{\nsim R}$$^{\nsim 0}$ & [0.000, 0.000]$^{\dagger}$ & 0.004$^{**}$ & [0.273, 0.339] & 1.000$^{\nsim R}$$^{\nsim 0}$ & [0.000, 0.000]$^{\dagger}$ \\
F$_{\beta=2}$ & 1.000$^{\nsim R}$$^{\nsim 0}$ & [0.000, 0.000]$^{\dagger}$ & 0.004$^{**}$ & [0.292, 0.381] & 0.500$^{\nsim R}$$^{\nsim 0}$ & [0.000, 0.007] \\
F1$_{\text{pa}}$ & 1.000$^{\nsim R}$$^{\nsim 0}$ & [0.000, 0.000]$^{\dagger}$ & 0.004$^{**}$ & [0.064, 0.101] & 1.000$^{\nsim R}$$^{\nsim 0}$ & [0.000, 0.000]$^{\dagger}$ \\
F1$_{\text{pa50\%}}$ & 1.000$^{\nsim R}$$^{\nsim 0}$ & [0.000, 0.000]$^{\dagger}$ & 0.008$^{**}$ & [0.141, 0.273] & 1.000$^{\nsim R}$$^{\nsim 0}$ & [0.000, 0.000]$^{\dagger}$ \\
F1$_{\text{w,10s}}$ & 0.016$^{*}$ & [0.005, 0.017] & 0.004$^{**}$ & [0.091, 0.159] & 0.008$^{**}$ & [0.042, 0.062] \\
F1$_{\text{w,30s}}$ & 0.004$^{**}$ & [0.019, 0.038] & 0.004$^{**}$ & [0.112, 0.167] & 0.008$^{**}$ & [0.089, 0.175] \\
F1$_{\text{w,5min}}$ & $<$0.001$^{***}$ & [0.305, 0.427] & 0.004$^{**}$ & [0.019, 0.050] & 0.008$^{**}$ & [0.231, 0.501] \\
F1$_{\text{w,20min}}$ & $<$0.001$^{***}$ & [0.613, 0.836] & 1.000$^{\nsim R}$ & [0.000, 0.000]$^{\ddagger}$ & 0.008$^{**}$ & [0.223, 0.583] \\
\bottomrule
\end{tabular}
\end{table}

\begin{itemize}
    \item \textbf{ADARP}: As suggested by the raw scores (\Cref{fig:bar_plot_comp}), standard and point-adjusted metrics show no significant difference between TimesFM and either baseline ($p=1.0^{\nsim R}$$^{\nsim 0}$), confirming no measurable performance. In contrast, the observed gaps for windowed metrics (F1$_w$) are statistically significant for 10s ($p=0.016$), 30s ($p=0.004$), 5min, and 20min windows ($p<0.001$), with confidence intervals indicating increasing effect sizes for larger windows.
    \item \textbf{ROAD}: For most metrics, the large performance gap observed relative to the random baseline is highly significant ($p=0.004$)\footnote{For ROAD, there are 9 subjects, so the smallest nonzero two-sided permutation $p$-value is $2\cdot1/2^9 \approx 0.004$. The factor of 2 is for both the all-positive and all-negative sign-flip permutations, which are equally extreme in absolute value.}, indicating that TimesFM consistently provides a meaningful improvement. However, for the 20min window, the metric becomes degenerate ($p=1.0$), confirming that the perfect scores seen previously are not discriminative.
    \item \textbf{Wrist Angel}: Similar to ADARP, standard and point-adjusted metrics show no significant difference from either baseline ($p=1.0^{\nsim R}$$^{\nsim 0}$ for F1, F$_{\beta=0.5}$, F1$_{\text{pa}}$, F1$_{\text{pa50\%}}$), confirming the lack of measurable performance. For F$_2$, the difference is also not significant ($p=0.5^{\nsim R}$$^{\nsim 0}$). However, the relatively higher F$_2$ score – which weights recall more strongly – suggests the model achieves high recall at the expense of precision, indicating a tendency to produce many false positives. In contrast, windowed metrics (F1$_w$) are statistically significant for all window sizes: 10s ($p=0.008^{**}$), 30s ($p=0.008^{**}$), 5min ($p=0.008^{**}$), and 20min ($p=0.008^{**}$), with confidence intervals indicating meaningful effect sizes.
\end{itemize}

\subsection{Full Results}
A complete table of all metric scores for each dataset and method is provided in Appendix~\ref{tab:performance_comparison}.

\section{Discussion and Conclusion}
Our results show that the proposed window-based F1$_w$ metric uncovers statistically significant model performance in imbalanced, real-world datasets, performance that is invisible to conventional (F1, F$_\beta$) and to adjusted metrics (F$_{pa}$, F$_{pa\%K}$). 

For the in-the-wild datasets ADARP and Wrist Angel, F1$_w$ was uniquely able to capture meaningful model performance, whereas conventional metrics yielded near-zero scores due to point-based annotations. In contrast, the ROAD dataset, with its long continuous stress segments, naturally resulted in high scores for all metrics, even for the random baseline model. This likely stems from the underlying data collection process: ROAD exhibits a much higher event prevalence (by a factor of $\sim 10^4$) compared to the in-the-wild datasets. These characteristics differ from in-the-wild annotation practices and effectively render the data "pre-adjusted" to highlight the stress signal. Consequently, tolerance-based and point-adjusted metrics can further amplify this inflation, and should be applied only with careful consideration of the dataset’s structure and annotation scheme. Alternatively, a comparison to a random baseline can provide such insight, in cases where prior domain knowledge may be absent.

Unlike conventional approaches that may require modifying or segmenting the ground-truth labels to enable evaluation, F1$_w$ allows direct, post-hoc assessment of predictions using the original annotations. In addition, the choice of the window size $w$ is directly interpretable as a time tolerance in the application domain. This demonstrates that F1$_w$ is robust in scenarios where annotation granularity and event timing are uncertain, and bypasses the need for artificial adjustment of ground-truth labels.

One limitation of our study is the use of the pre-trained TimesFM in an event prediction setting. TimesFM is pre-trained to forecast a target signal given the previous history of the target signal, and in our experiments, the stress labels themselves are used as that target signal. This setup allows the model to use past stress labels as context for forecasting future events, which can result in artificially high probabilities around annotated events. In a real-world deployment, such explicit stress labels may not be available as input, and thus may create a setting that is not realistic for some applications of stress event detection. However, our focus has been an investigation of F-metrics for evaluation of event detection, and for this purpose, the modeling is comparable across all metrics. For future event prediction, it may be more realistic for the model to rely only on physiological signals and other covariates to inform its forecast of the target signal, something which the current version of TimesFM does not support. An additional limitation is the limited public availability of in-the-wild datasets with stress annotations, restricting broader validation and reproducibility. 

The choice of window size in F1$_w$ is critical. Overly broad windows can overestimate performance, while overly narrow windows may revert to the limitations of point-based metrics. Selecting an appropriate window size requires domain expertise and careful consideration of the application context. On the other hand, using the windowed F-metric has the advantage that it is directly related to domain knowledge as a temporal tolerance.

For simplicity, we have not incorporated a point-precision adjustment into F1$_w$, though our framework allows for flexible windowing in precision and recall calculations. Using a stricter window for precision, for example, would make the metric behave more like point-precision, allowing for nuanced control over the balance between recall and precision in the F1 score.

In summary, the window-based F1 metric offers a robust and practical approach for evaluating event detection in time series, particularly in healthcare contexts characterized by temporal ambiguity and annotation variability. By enabling post-hoc, domain-adaptive assessment, F1$_w$ supports more meaningful benchmarking and guides the development of reliable interventions in real-world settings.
\paragraph{Ethical Considerations}
With F1$_w$, post-hoc evaluation uses the original point annotations, avoiding artificial label adjustments that can inflate results, thereby improving cross-study consistency and reproducibility. Although the use cases (stress related to AUD, OCD, or driving) are benevolent, sensitive physiological and mental health–related predictions could be misused (e.g., intrusive monitoring or stigmatization).

\section{Data and Code Availability}
The code for this project is openly available on GitLab (\url{https://lab.compute.dtu.dk/harsk/window-f1-metric}). The ADARP dataset is available at \url{https://zenodo.org/records/6640290}, and the ROAD dataset is available at \url{https://www.media.mit.edu/tools/affectiveroad/}. The Wrist Angel dataset is not publicly available due to privacy concerns; however, anonymized labels and predictions are included in the code repository.

\section{Acknowledgments}
\textbf{Funding}: This work was partially supported by the \textit{Innovation Fund Denmark} project "Personas and Behaviors of AI for Social Skills and Mental Health (PebAI4MH)," which provided salary support.

\bibliographystyle{unsrtnat}  
\bibliography{references}

\appendix
\section{Model Setup Illustration}
\begin{figure}
    \centering
    \includegraphics[width=0.98\linewidth]{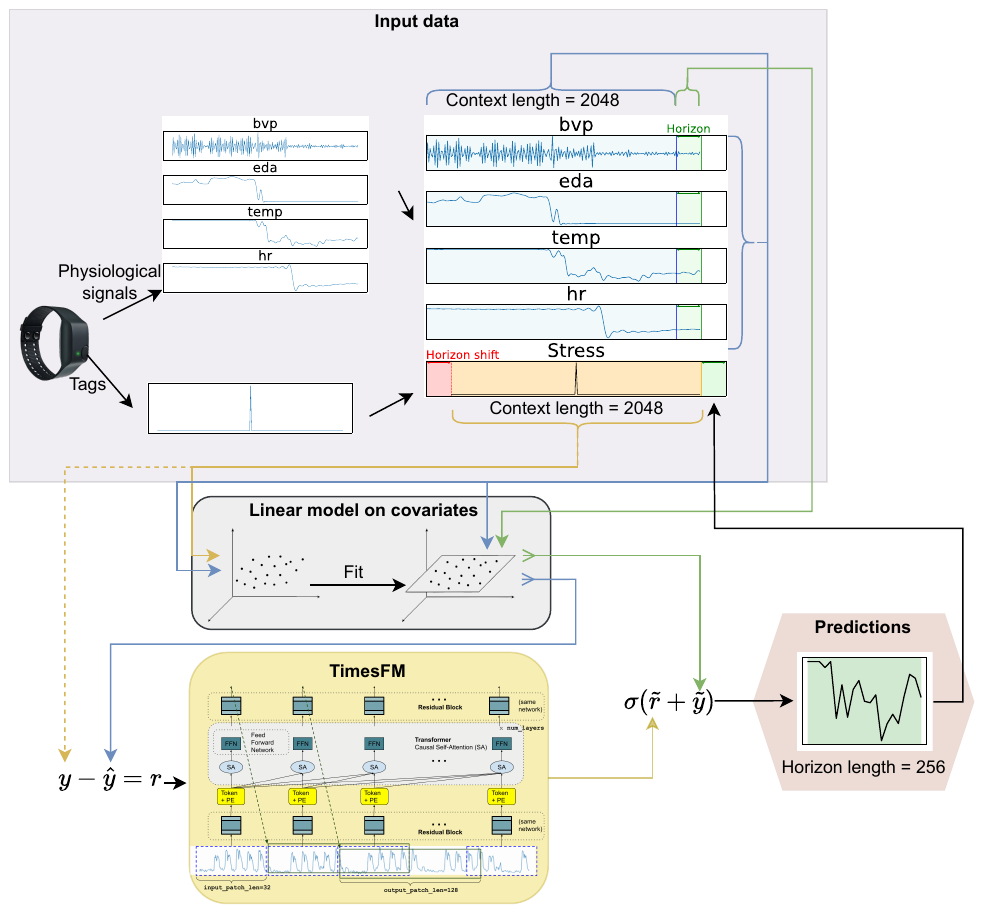}
    \caption{Illustration of the modeling setup. The TimesFM architecture illustration in the yellow box is from \cite{GoogleTimesFM}.}
    \label{fig:illustration}
\end{figure}
Model setup and data input are illustrated in \cref{fig:illustration}. Note that the stress signal lags behind the covariates. TimesFM incorporates covariates that are available at prediction time, aligning them with the forecast horizon. By shifting the stress signal forward by the length of the horizon window, the model uses covariates to predict future stress values, rather than relying on information about the current stress measurement.

\section{Hardware Specifications and Inference Runtimes}
\label{sec:runtime_details}

Inference runtimes and hardware configurations varied across datasets due to differences in computational resources and data sensitivity.

\newlist{hwdesc}{description}{1}
\setlist[hwdesc]{style=unboxed,font=\normalfont,labelsep=0.5em,itemsep=0pt,parsep=0pt,topsep=2pt}

\subsection{ADARP and ROAD}
\begin{hwdesc}
  \item[System] Linux
  \item[GPU] NVIDIA RTX A5000 (Ampere, 24 GB)
  \item[CPU] AMD EPYC 7252 (8c/16t, 3.1 GHz)
  \item[RAM] 128 GB
\end{hwdesc}

\subsection{Wrist Angel}
\begin{hwdesc}
  \item[System] Windows
  \item[GPU] Not supported on Windows for current JAX (\cite{jax})
  \item[CPU] Intel Core i9-10900X (3.7 GHz)
  \item[RAM] 32 GB
\end{hwdesc}

\begin{table}[h!]
\centering
\caption{Prediction runtimes and typical per-signal prediciton times for each dataset and system.}\label{tab:hardware_specs}
\begin{tabular}{l|c|c|c}
Dataset & System & Tot. Runtime (h) & Prediction Time (s) \\
\midrule 
ADARP & GPU & 2.5 & \num{0.002} \\
ROAD & GPU & $<$0.1 & \num{0.00047} \\
Wrist Angel & CPU & 24 & 0.22\\
\end{tabular}
\end{table}

The longer total runtime for Wrist Angel reflects CPU-only computation due to Windows incompatibility with TImesFM jax-cuda backend.

\section{Thresholds}
\begin{figure}[H]
    \centering
    \includegraphics[width=0.75\linewidth]{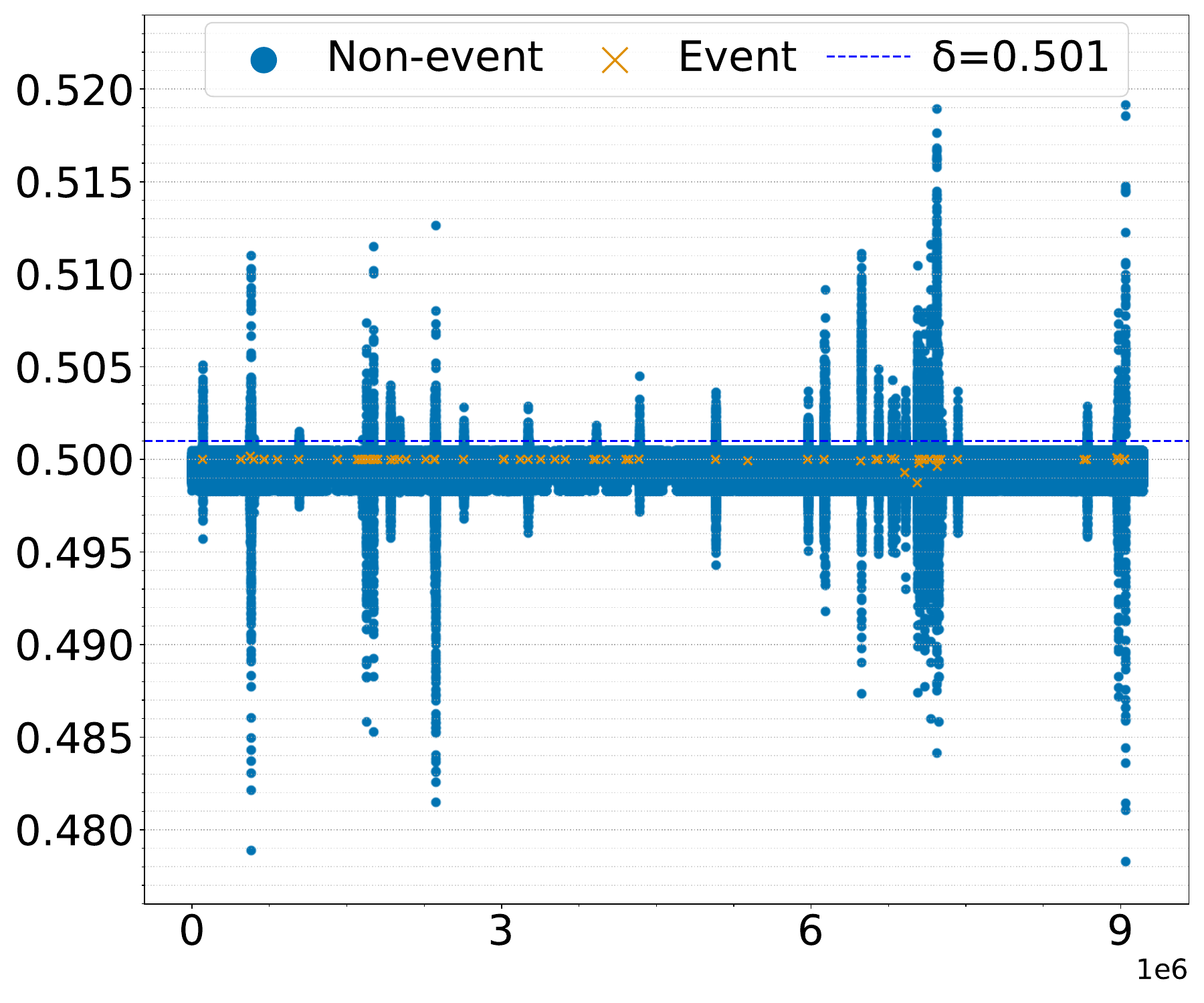}
    \caption{ADARP predicted probabilities. Context = 2048 (512 seconds, 8.5 minutes) horizon = 256 (64 seconds).}
    \label{fig:ADARP_val_prob_dist}
\end{figure}

\begin{figure}[H]
    \centering
    \includegraphics[width=0.75\linewidth]{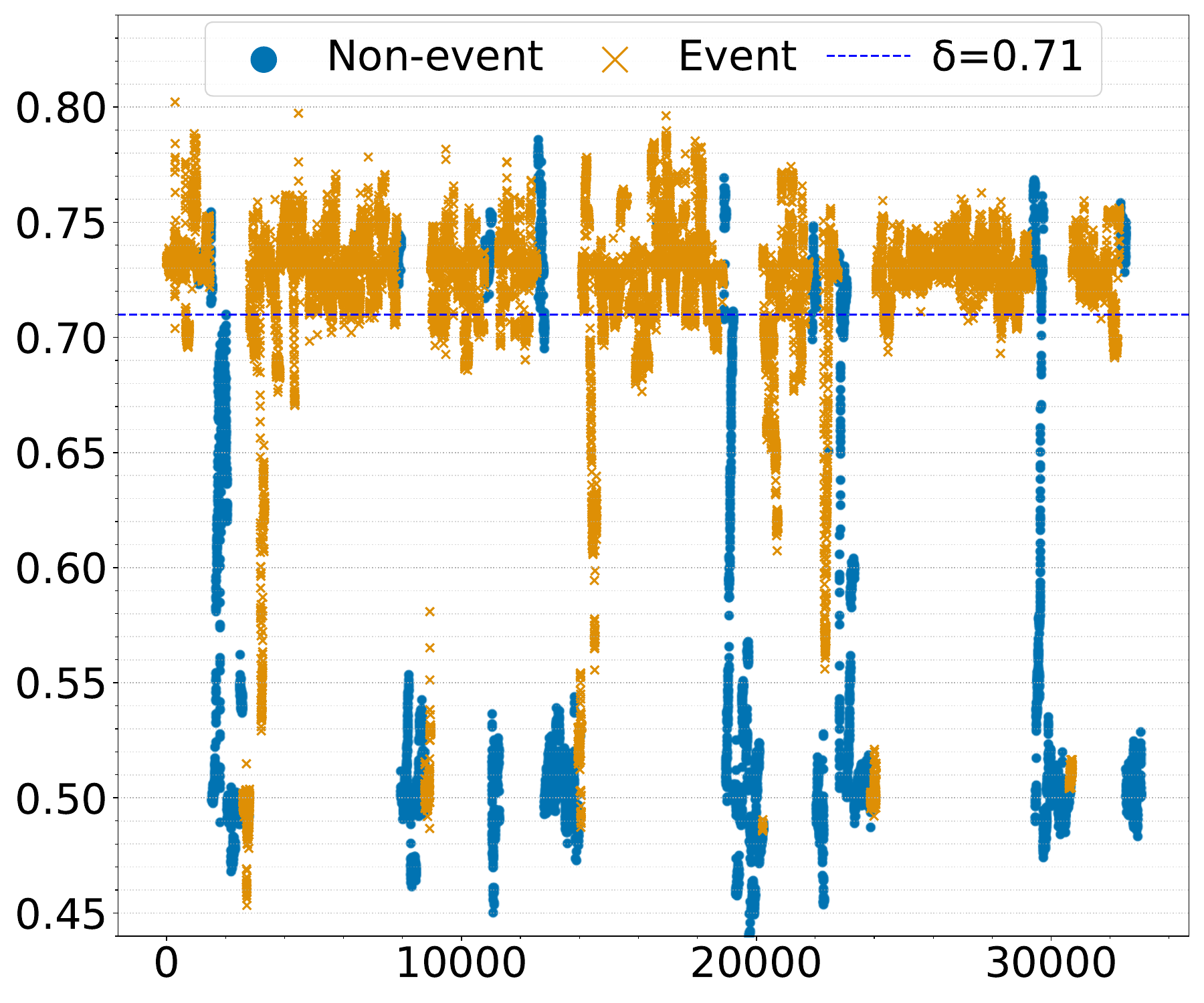}
    \caption{ROAD predicted probabilities. Context = 2048 (512 seconds, 8.5 minutes) horizon = 256 (64 seconds).}
    \label{fig:ROAD_val_prob_dist}
\end{figure}

\begin{figure}[H]
    \centering
    \includegraphics[width=0.75\linewidth]{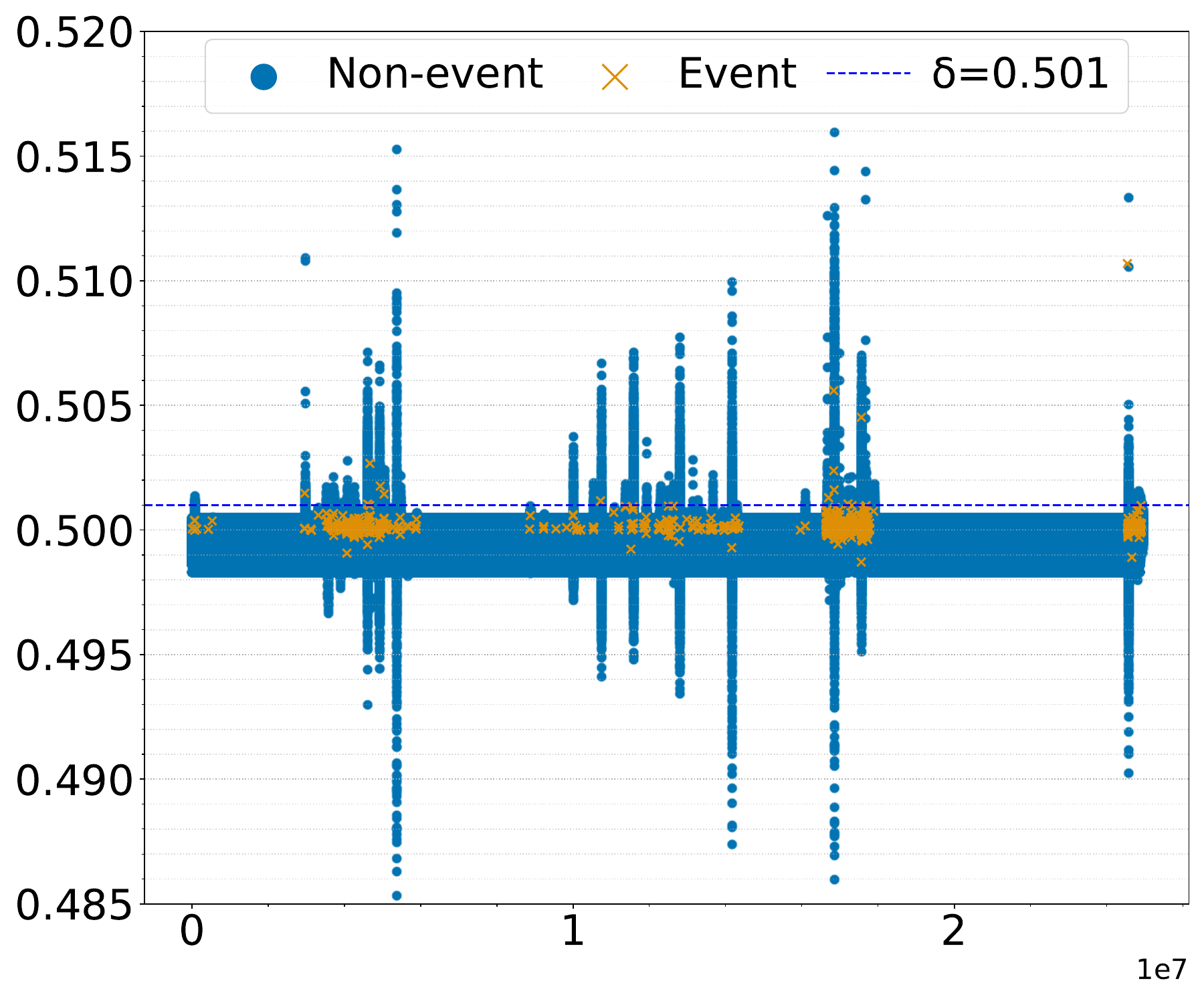}
    \caption{Wrist Angel predicted probabilities. Context = 2048 (512 seconds, 8.5 minutes) horizon = 256 (64 seconds).}
    \label{fig:WA_val_prob_dist}
\end{figure}

Prediction probabilities and labels, along with the chosen thresholds are visualized in Figures \ref{fig:ADARP_val_prob_dist}, \ref{fig:ROAD_val_prob_dist}, and \ref{fig:WA_val_prob_dist}.

\section{F-Metric Results}
The full results for all F-metrics are reported in Table \ref{tab:performance_comparison}.

\begin{table}[ht]
\centering
\footnotesize
\caption{Performance comparison across datasets. Metrics are grouped by hyperparameter: Point-Adjusted ($K$) and windowed ($w$).}
\label{tab:performance_comparison}
\begin{tabular}{@{}l*{3}{c}@{}}
\toprule
Metric & \textbf{ADARP} & \textbf{ROAD} & \textbf{WA} \\
\midrule
\multicolumn{4}{@{}l@{}}{\textbf{Standard Metrics}} \\
\addlinespace[0.5ex]
\hspace{1em}F1 & 0.0001 & 0.8756 & 0.0013 \\
\hspace{1em}Precision & 0.0000 & 0.9022 & 0.0007 \\
\hspace{1em}Recall & 0.0032 & 0.8506 & 0.0098 \\
\addlinespace[1ex]
\multicolumn{4}{@{}l@{}}{\textbf{F-beta Variants}} \\
\addlinespace[0.5ex]
\hspace{1em}F$_{0.5}$ & 0.0001 & 0.8914 & 0.0009 \\
\hspace{1em}F$_{2}$ & 0.0002 & 0.8604 & 0.0028 \\
\addlinespace[1ex]
\multicolumn{4}{@{}l@{}}{\textbf{Point-Adjusted Metrics}} \\
\addlinespace[0.5ex]
\multicolumn{4}{@{}l@{}}{\hspace{1em}K = 0} \\
\addlinespace[0.3ex]
\hspace{2em}F1$_{\text{PA}}$ & 0.0001 & 0.9443 & 0.0013 \\
\hspace{2em}Prec$_{\text{PA}}$ & 0.0000 & 0.9138 & 0.0007 \\
\hspace{2em}Rec$_{\text{PA}}$ & 0.0032 & 0.9770 & 0.0098 \\
\addlinespace[0.5ex]
\multicolumn{4}{@{}l@{}}{\hspace{1em}K = 0.25} \\
\addlinespace[0.3ex]
\hspace{2em}F1$_{\text{PA,25\%}}$ & 0.0001 & 0.9340 & 0.0013 \\
\hspace{2em}Prec$_{\text{PA,25\%}}$ & 0.0000 & 0.9121 & 0.0007 \\
\hspace{2em}Rec$_{\text{PA,25\%}}$ & 0.0032 & 0.9570 & 0.0098 \\
\addlinespace[0.5ex]
\multicolumn{4}{@{}l@{}}{\hspace{1em}K = 0.5} \\
\addlinespace[0.3ex]
\hspace{2em}F1$_{\text{PA,50\%}}$ & 0.0001 & 0.9182 & 0.0013 \\
\hspace{2em}Prec$_{\text{PA,50\%}}$ & 0.0000 & 0.9096 & 0.0007 \\
\hspace{2em}Rec$_{\text{PA,50\%}}$ & 0.0032 & 0.9271 & 0.0098 \\
\addlinespace[0.5ex]
\multicolumn{4}{@{}l@{}}{\hspace{1em}K = 0.75} \\
\addlinespace[0.3ex]
\hspace{2em}F1$_{\text{PA,75\%}}$ & 0.0001 & 0.8929 & 0.0013 \\
\hspace{2em}Prec$_{\text{PA,75\%}}$ & 0.0000 & 0.9053 & 0.0007 \\
\hspace{2em}Rec$_{\text{PA,75\%}}$ & 0.0032 & 0.8808 & 0.0098 \\
\addlinespace[1ex]
\multicolumn{4}{@{}l@{}}{\textbf{Windowed Metrics}} \\
\addlinespace[0.5ex]
\multicolumn{4}{@{}l@{}}{\hspace{1em}$w$ = 10s} \\
\addlinespace[0.3ex]
\hspace{2em}F1$_{w,\text{10s}}$ & 0.0104 & 0.9029 & 0.0561 \\
\hspace{2em}Prec$_{w,\text{10s}}$ & 0.0057 & 0.9197 & 0.0450 \\
\hspace{2em}Rec$_{w,\text{10s}}$ & 0.0613 & 0.8867 & 0.0744 \\
\addlinespace[0.5ex]
\multicolumn{4}{@{}l@{}}{\hspace{1em}$w$ = 30s} \\
\addlinespace[0.3ex]
\hspace{2em}F1$_{w,\text{30s}}$ & 0.0352 & 0.9386 & 0.1269 \\
\hspace{2em}Prec$_{w,\text{30s}}$ & 0.0203 & 0.9451 & 0.1221 \\
\hspace{2em}Rec$_{w,\text{30s}}$ & 0.1323 & 0.9322 & 0.1321 \\
\addlinespace[0.5ex]
\multicolumn{4}{@{}l@{}}{\hspace{1em}$w$ = 1min} \\
\addlinespace[0.3ex]
\hspace{2em}F1$_{w,\text{1min}}$ & 0.0796 & 0.9724 & 0.1788 \\
\hspace{2em}Prec$_{w,\text{1min}}$ & 0.0493 & 0.9710 & 0.2153 \\
\hspace{2em}Rec$_{w,\text{1min}}$ & 0.2065 & 0.9738 & 0.1528 \\
\addlinespace[0.5ex]
\multicolumn{4}{@{}l@{}}{\hspace{1em}$w$ = 5min} \\
\addlinespace[0.3ex]
\hspace{2em}F1$_{w,\text{5min}}$ & 0.4019 & 0.9975 & 0.3088 \\
\hspace{2em}Prec$_{w,\text{5min}}$ & 0.3317 & 0.9997 & 0.6691 \\
\hspace{2em}Rec$_{w,\text{5min}}$ & 0.5097 & 0.9954 & 0.2007 \\
\addlinespace[0.5ex]
\multicolumn{4}{@{}l@{}}{\hspace{1em}$w$ = 20min} \\
\addlinespace[0.3ex]
\hspace{2em}F1$_{w,\text{20min}}$ & 0.7984 & 1.0000 & 0.4586 \\
\hspace{2em}Prec$_{w,\text{20min}}$ & 1.0000 & 1.0000 & 0.9991 \\
\hspace{2em}Rec$_{w,\text{20min}}$ & 0.6645 & 1.0000 & 0.2976 \\
\addlinespace[0.5ex]
\multicolumn{4}{@{}l@{}}{\hspace{1em}$w$ = 60min} \\
\addlinespace[0.3ex]
\hspace{2em}F1$_{w,\text{60min}}$ & 0.8100 & 1.0000 & 0.6038 \\
\hspace{2em}Prec$_{w,\text{60min}}$ & 1.0000 & 1.0000 & 0.9994 \\
\hspace{2em}Rec$_{w,\text{60min}}$ & 0.6806 & 1.0000 & 0.4325 \\
\bottomrule
\end{tabular}
\end{table}

\end{document}